\documentclass{article}
\usepackage{spconf,amsmath,graphicx}
\usepackage{multirow,diagbox}
\usepackage{footnote}
\makesavenoteenv{tabular}
\makesavenoteenv{table}

\title{Prediction-Adaptation-Correction Recurrent Neural Networks \\for Low-Resource Language Speech Recognition}
\name{Yu Zhang$^{1}$, Ekapol Chuangsuwanich$^{1}$, James Glass$^{1}$,  Dong Yu$^{2}$}
\address{
\begin{tabular}{cl}
    $^1$MIT Computer Science and Artificial Intelligence Laboratory & $^2$Microsoft Research\\
   {\small \tt \{yzhang87, ekapolc, glass\}@mit.edu} & {\small \tt dongyu@microsoft.com}
    \thanks{This work was supported in part by the Intelligence Advanced Research Projects Activity (IARPA) via Department of Defense US Army Research Laboratory contract number W911NF-12-C-0013. The U.S. Government is authorized to reproduce and distribute reprints for Governmental purposes notwithstanding any copyright annotation thereon. Disclaimer: The views and conclusions contained herein are those of the authors and should not be interpreted as necessarily representing the official policies or endorsements, either expressed or implied, of IARPA, DoD/ARL, or the U.S. Government.}
\end{tabular}
 }
\begin{document}
%
\maketitle
\vspace*{-0.5in}
\begin{abstract}
In this paper, we investigate the use of prediction-adaptation-correction recurrent neural networks (PAC-RNNs) for low-resource speech recognition. A PAC-RNN is comprised of a pair of neural networks in which a {\it correction} network uses auxiliary information given by a {\it prediction} network to help estimate the state probability. The information from the correction network is also used by the prediction network in a recurrent loop. Our model outperforms other state-of-the-art neural networks (DNNs, LSTMs) on IARPA-Babel tasks. Moreover, transfer learning from a language that is similar to the target language can help improve performance further.
\end{abstract}
\section{Introduction}
\label{sec:intro}
The behavior of prediction, adaptation, and correction is widely observed in human speech recognition \cite{feedbackRef}. For example, listeners may guess what you will say next and wait to confirm their guess. They may adjust their listening effort by predicting the speaking rate and noise condition based on current information, or predict and adjust a letter to sound mapping based on the talker's pronunciations.

Previously~\cite{YuPACRNN}, we proposed the prediction-adaptation-correction RNN (PAC-RNN) which tries to emulate some of these mechanisms by using two DNNs; a {\it prediction} DNN that predicts the next phoneme, and a {\it correction} DNN that estimates the current state probability based on both the current frame and the hypothesis from the prediction DNN. The model showed promising results on TIMIT, but it was unclear whether a similar gain could be achieved on larger ASR tasks where the prediction information might already be incorporated by the language models. Here, we successfully apply the PAC-RNN to LVCSR on several low-resource languages that are currently being used in the IARPA-Babel program.

In addition, we study the effect of transfer learning for recurrent architectures. Recurrent networks such as LSTMs~\cite{BLSTMTIMIT, BLSTMAM} are known to require a large amount of training data in order to perform well \cite{LSTM-Sak+2014}. For the IARPA-Babel tasks, multiple groups have incorporated multilingual training in order to alleviate data limitation issues. One popular approach is multi-task training using DNNs. In a multi-task setup, a single DNN is trained to generate outputs for multiple languages with some tied parameters. This approach has been used for robust feature extraction via bottleneck (BN) features~\cite{aachen,LID,BUTadapt,CMUMultiGPU}, or for classifiers in hybrid DNN-HMM approaches ~\cite{MicrosoftDNN,mcgil}. In \cite{BUTSBNhybrid}, Karafi\'{a}t et al.~found that using CMLLR transformed BN features as inputs to a hybrid DNN could further improve ASR performance. However, we believe none of this research has investigated recurrent networks for low resource languages in a multilingual scenario.

The work presented here is an extension of \cite{YuPACRNN} based on our multilingual framework in \cite{LID}. We first extract BN features using multilingual networks to train different hybrid neural network architectures. Experiments show that the LSTMs outperform DNNs, and that the PAC-RNN provides the biggest gains for this task. Additional improvements are observed when the models are adapted from networks trained on languages that are most similar to the target language.

The rest of the paper is organized as follows. In Section \ref{sec:PAC-RNN}, we review the PAC-RNN model, and describe an enhanced version that incorporates an LSTM. In Section \ref{sec:sbn}, we describe our multilingual system and how it is used with the PAC-RNN. We explain our experiments results in Section \ref{sec:experiments}.

\section{Prediction-Adaptation-Correction Recurrent Neural Networks}
\label{sec:PAC-RNN}
\subsection{Model structure and training}
\begin{figure*}[t]
\centering
\includegraphics[width=0.75\textwidth]{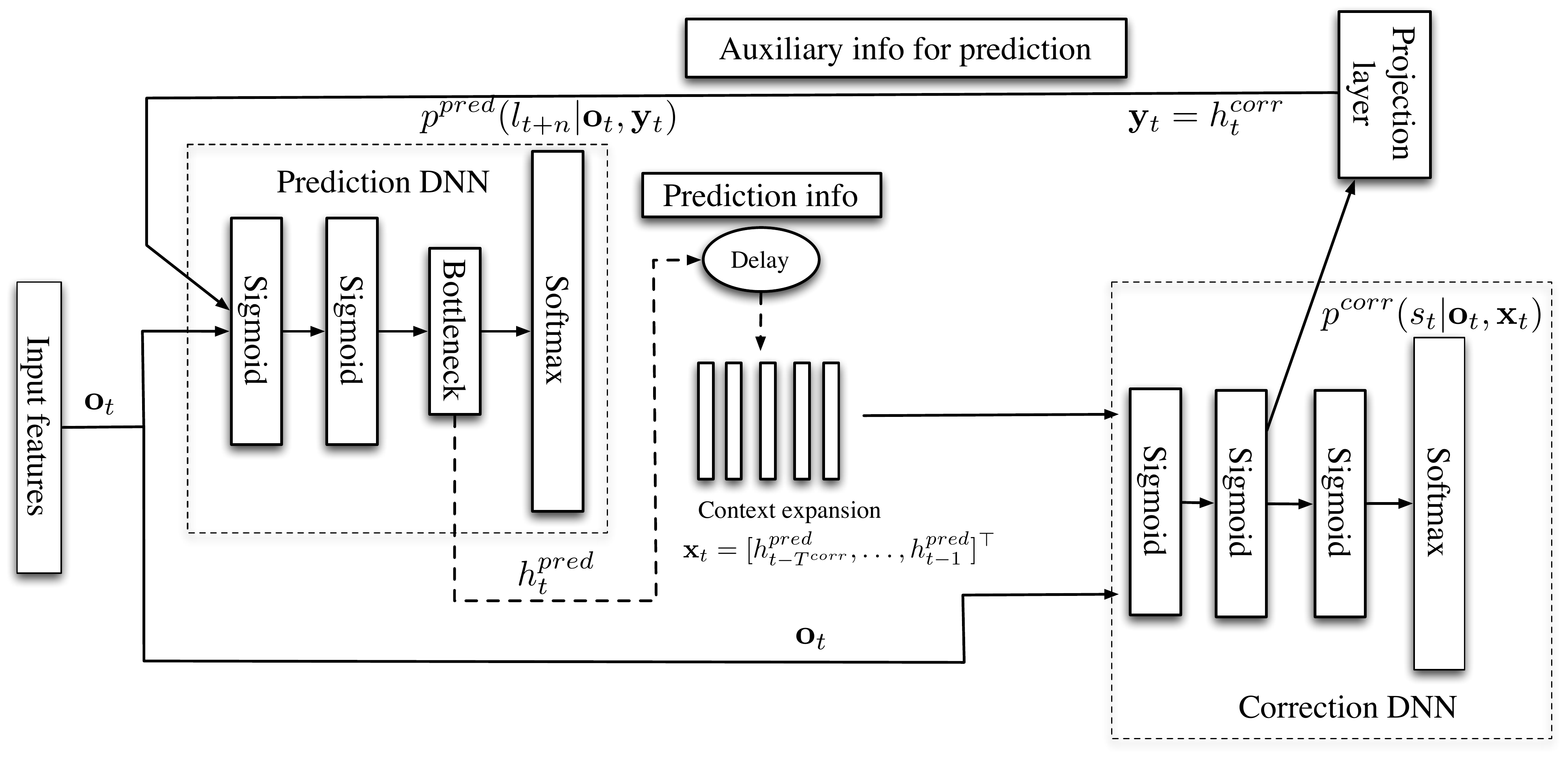}
\caption{The Structure of the PAC-RNN-DNN}
\label{fig:PCRNN}
\end{figure*}
The PAC-RNN used in this work follows our previous work in \cite{YuPACRNN}.
Fig.~\ref{fig:PCRNN} illustrates the structure of the PAC-RNN studied in this paper. The main components of the model are a {\it correction} DNN and a {\it prediction} DNN. The correction DNN estimates the state posterior probability $p^{corr}(s_t|\mathbf{o}_t, \mathbf{x}_t)$ given $\mathbf{o}_t$, the observation feature vector, and $\mathbf{x}_t$, the information from the prediction DNN, at time $t$. The prediction DNN predicts future target information. Note that since  $\mathbf{y}_t$, the information from the correction DNN, depends on $\mathbf{x}_t$, the information from the prediction DNN, and vice versa, a recurrent loop is formed.

The information from the prediction DNN, $\mathbf{x}_t$, is from a bottleneck hidden layer output value $h^{pred}_{t-1}$. To exploit additional previous predictions, we stack multiple hidden layer values as 
\begin{equation}
    \mathbf{x}_t = [h^{pred}_{t-T^{corr}}, . . . , h^{pred}_{t-1}]^T,
\end{equation}
where $T^{corr}$ is the contextual window size used by the correction DNN and is set to 10 in our study. Similarly, we can stack multiple frames to form $\mathbf{y}_t$, the information from the correction DNN, as
\begin{equation}
    \mathbf{y}_t = [h^{corr}_{t-T^{pred}-1}, . . . , h^{corr}_{t}]^T,
\end{equation}
where $T^{pred}$ is the contextual window size used by the prediction DNN and is set to 1 in our study. In addition, in the specific example shown in Fig. \ref{fig:PCRNN}, the hidden layer output $h^{corr}_{t}$ is projected to a lower dimension before it is fed into the prediction DNN.

To train the PAC-RNN, we need to provide supervision information to both the prediction and  correction DNNs. As we have mentioned, the correction DNN estimates the state posterior probability, and thus the state label, so that the frame cross-entropy (CE) criterion can be used. For the prediction DNN, we follow~\cite{YuPACRNN}, and use the phoneme label for the prediction targets.

The PAC-RNN training problem is a multi-task learning problem. The two training objectives can be combined into a single one as 
\begin{equation}\label{eq:criterion}
    J=\sum_{t=1}^T (\alpha*\ln p^{corr}(s_t|\mathbf{o}_t, \mathbf{x}_t) + (1-\alpha)*\ln p^{pred}(l_{t+n}|\mathbf{o}_t, \mathbf{y}_t)),
\end{equation}
where $\alpha$ is the interpolation weight, and is set to 0.8 in our study unless otherwise stated, and $T$ is the total number of frames in the training utterance. Note that in a standard PAC-RNN as described here, both the correction model and prediction model are DNNs. From this point onwards we will call this particular setup, the PAC-RNN-DNN.

\subsection{PAC-RNN-LSTM}
LSTMs have improved speech recognition accuracy on many tasks over DNNs \cite{BLSTMTIMIT, BLSTMAM, LSTM-Sak+2014}. To further enhance the PAC-RNN model, we use an LSTM to replace the DNN used in the correction model. The input of this LSTM is the acoustic feature $\mathbf{o}_t$ concatenated with the information from prediction model, $\mathbf{x}_t$. The prediction model can also be an LSTM but we did not observe performance gain on the experiments. To keep it simple, we use the same DNN prediction model as~\cite{YuPACRNN}. 

\section{Stacked bottleneck architecture}
\label{sec:sbn}

\subsection{Stacked bottleneck (SBN) features}
The BN features used in this work follow our previous work in~\cite{LrSBN}. An SBN is a hierarchical architecture realized as a concatenation of two DNNs, each with its own bottleneck layer. The outputs from the BN layer in the first DNN are used as the input features for the second DNN,  whose outputs at the BN layer are then used as the final features for standard GMM-HMM training. 



\subsection{Bottleneck-CMLLR features in a hybrid system}
In \cite{BUTSBNhybrid}, the authors proposed a DNN hybrid system that used the first stage BN features with speaker adaptation (BN-CMLLR). This system yielded the best results for the Babel evaluation that year. In this work, we follow a similar approach by replacing the DNN with recurrent architectures (LSTM or PAC-RNN). The BN-CMLLR features were taken from a network trained in a multilingual fashion and adapted to the target language. For the DNN and PAC-RNN, these features were stacked in context of $31 (\pm 15)$ frames and downsampled by a factor of $5$. Following \cite{LSTM-Sak+2014}, no context expansion is used for the LSTM. The output state label is also delayed by $5$ to utilize the information from the future. 

\subsection{Multilingual training and adaptation of SBN features}
\label{subsec:multsbn}



The multilingual training of the SBN follows \cite{MultiOne} where all the DNN targets from each language are pooled together, each with its own softmax layer. Adapting the multilingual SBN to a target language can be done by performing additional fine-tuning steps on each DNN sequentially using the data from the target language. Our previous work~\cite{LID} shows that using just the language closest to the target language from the pool of source languages to train the second DNN can serve as a better initialization model than the multilingual second DNN. The closest language can be identified from just the acoustic data by training a Language Identification (LID) system.

\begin{figure}
  \centering
  \centerline{\includegraphics[width=0.45\textwidth]{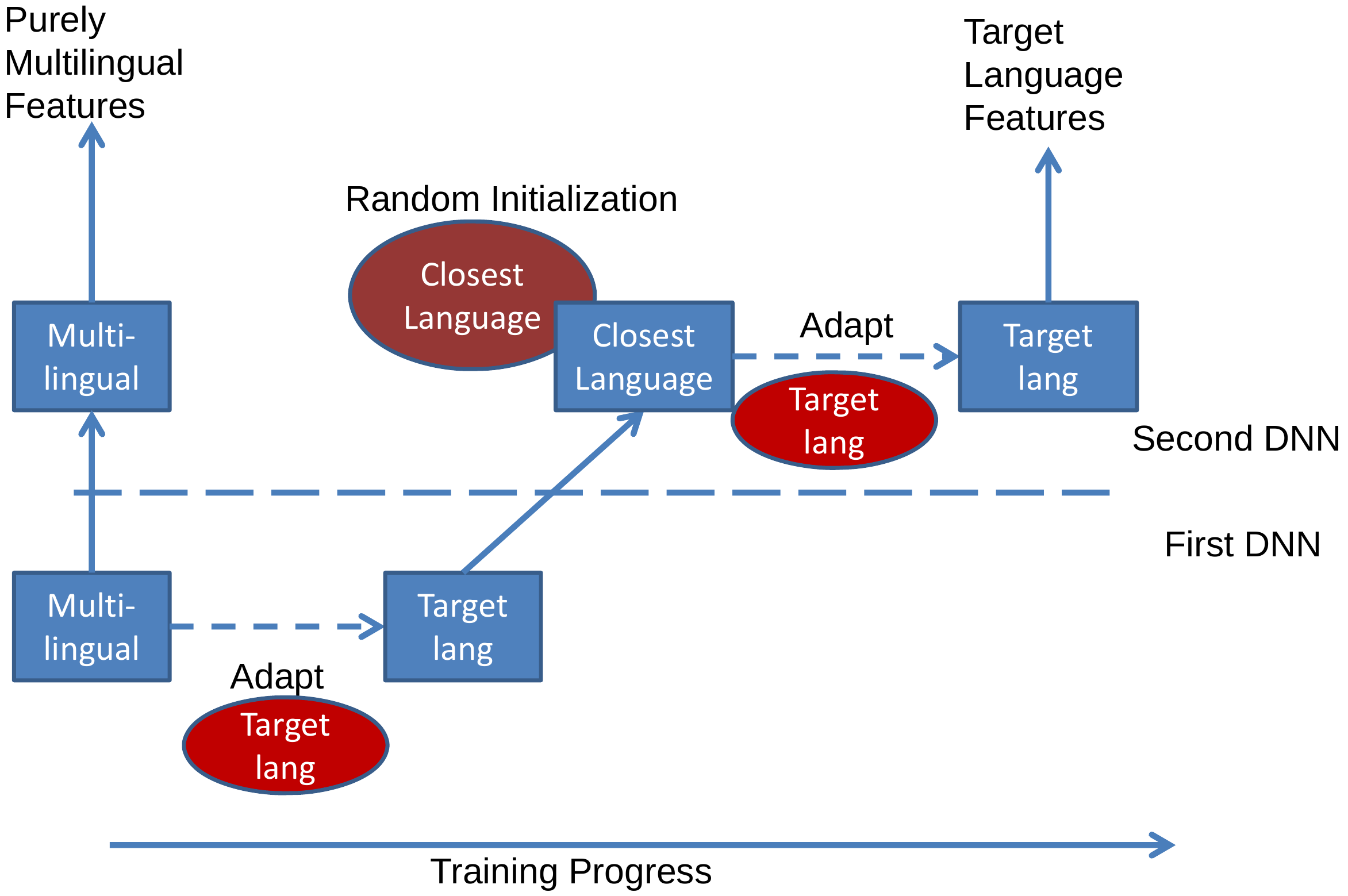}}
  \vspace{-10pt}
  \caption{\small{Steps to adapt a multilingual SBN to a target language via the closest language selected via LID.}}
  \label{fig:closestlang}
\end{figure}

A flowchart of how a LID-based multilingual system can be trained is shown in Fig.~\ref{fig:closestlang}. We start by adapting the first DNN with data from the target language. Instead of using the second multilingual DNN to initialize, we train the second DNN from random initialization using the closest language's data and output targets. After the DNN converges, we then do a final adaptation to the target language.

\subsection{Multilingual training of BN-hybrid system}
The input of the hybrid system (DNN, LSTM or PAC-RNN) is the same as the second DNN in the SBN system. During the adaptation stage, the softmax is replaced by the target language state labels (phone labels for the PAC-RNN prediction model) with random initialization while the hidden layers are initialized from the DNN, LSTM or PAC-RNN which is trained using the closest language. 

\section{Experiments}
\label{sec:experiments}

\subsection{IARPA-Babel corpus}

The IARPA-Babel program focuses on ASR and spoken term detection on low-resource languages~\cite{Babel}. The goal of the program is to reduce the amount of time needed to develop ASR and spoken term detection capabilities in a new language. The data from the Babel program consists of collections of speech from a growing list of languages. The project is on its fourth year. For this work we will consider the Full pack (60-80 hours of training data) of the 11 languages released in the first two years as source languages, while the languages in the third year will be the target languages ~\cite{FrameSelection}. Some languages also contain a mixture of microphone data recorded at 48kHz in both train and test utterances. For the purpose of this paper, we downsampled all the wideband data to 8kHz and treated it the same way as the rest of the recordings. For the target languages, we will focus on the Very Limited Language Pack (VLLP) condition which includes only 3 hours of transcribed training data. This condition excludes any use of human generated pronunciation dictionary. Unlike in the previous two years of the program, usage of web data is permitted for language modeling and vocabulary expansion. 

\subsection{Recognition system}
For each language, we used tied-state triphone CD-HMMs, with ~2500 states and 18 Gaussian components per state. Grapheme-based dictionaries were used for the target languages. Note that for IARPA-Babel languages, the difference between phonetic and graphemic systems in WER are often less than 1\% \cite{LIMSI,CUEDgrapheme}. All the output targets were from CD states. To train the multilingual SBN, we kept only the SIL frames that appear 5 frames before and after actual speech. This reduced the total amount of frames for the multilingual DNN to around 520 hours. We observed no loss in accuracy from doing so, and it also reduced the training time significantly. Discriminative training was done on the CD-HMMs using the Minimum Bayes risk (MBR) criterion~\cite{MBR}. The web data was cleaned and filtered using techniques described in ~\cite{bbnweb}. For language modeling, n-gram LMs were created from training data transcripts and the web data. The LMs were then combined using weighted interpolation. The vocabulary included words that appeared in the training transcripts augmented with the top 30k most frequent words from the web.  We chose 30k words by looking at the rate of OOV reduction as we augmented the train vocabulary with frequent words from the web. We report results on the 10-hour development set.

We consider two baseline hybrid systems: a DNN with three 1024-unit hidden layers, and a stacked LSTM with three layers each containing 512 cells. No gains were observed by further increasing the model size of these baseline systems.

In the PAC-RNN model, the prediction DNN has a 2048-unit hidden layer and a 80-unit bottleneck layer. For the correction model, we have two systems: a DNN with two 2048-unit hidden layers or an LSTM with 1024 memory cells. The  correction model's projection layer contains 500 units.

All models are randomly initialized without either generative or discriminative pretaining. No momentum is used for the first epoch and a momentum of $0.9$ is used for all the subsequent epochs. We have found that turning off the momentum for the first epoch helps to improve the performance of the final model. To train the DNN, a learning rate of $0.1$ per mini-batch is used for the first epoch. The learning rate is increased to $1.0$ at the second epoch, after which it is kept the same until the development set training criterion no longer improves, under which condition the learning rate is halved. A similar schedule is used to train the LSTMs and PAC-RNNs except that all the learning rates are reduced to $1/10$ of that used in the DNN training. 

We implemented the hybrid models using the computational network toolkit (CNTK)~\cite{CNTK}. The truncated back-propagation-through-time (BPTT)~\cite{BPTT} is used to update the model parameters and each utterance is segmented into multiple chunks. To speed up the training, we process multiple utterances simultaneously as a batch. We have found that this reduces training time and improves the quality of the final model. In this study each BPTT segment contains 20 frames and we process 20 utterances simultaneously. For decoding, we fed the posteriors generated by CNTK into the Kaldi ASR toolkit~\cite{Kaldi}, which then generates the recognition results.

\subsection{PAC-RNN results with BN features}
Table~\ref{tab:results} summarizes the WERs achieved with different models evaluated in this study. The first three rows are the results from SBN systems. Both the multilingual and the closest language systems are adapted to the target language for the whole stacked network. For the hybrid systems, the input is the BN features extracted from the first DNN of the adapted multilingual SBN. 

The DNN hybrid system outperforms the multilingual SBN but is very similar to the closest language system. The LSTM improves upon the DNN by around 1\%. The PAC-RNN-DNN outperforms LSTM by another percent across all languages. By simply replacing the correction model with a single layer LSTM, we observe even further improvements.

\begin{table}[!ht]
\centering
\begin{tabular}{|l|c|c|c|}
\hline
Target language & Cebuano & Kurmanji & Swahili \\
\hline
Closest language & Tagalog & Turkish & Zulu \\
\hline
\multicolumn{4}{|l|}{ SBN models }\\
\hline
 Monolingual       & 73.5  & 86.2 & 65.8 \\
 Adapted multilingual  & 65.0  & 75.5 & 54.9 \\
 Closest language    & 63.7  & 75.0 & 54.2 \\
\hline
\hline
\multicolumn{4}{|l|}{ Hybrid models }\\
\hline
 DNN                   & 63.9  & 74.9 & 54.0 \\
 LSTM                  & 63.0  & 74.0 & 53.0 \\
 PACRNN-DNN          & 62.1  & 72.9 & 52.1 \\
 PACRNN-LSTM         & 60.6  & 72.5 & 51.4 \\
\hline
\hline
\multicolumn{4}{|l|}{Hybrid models with closest language initialization} \\
\hline
 DNN                      & 62.7  & 73.1 & 52.4 \\
 LSTM                     & 61.3  & 72.5 & 52.2 \\
 PAC-RNN-DNN             & 60.8  & 71.8 & 51.6 \\
 PAC-RNN-LSTM            & 59.7  & 71.4 & 50.4 \\
\hline
\end{tabular}
\caption{{WER (\%) results for each ASR system. }}
\label{tab:results}
\end{table}

\subsection{Effect of transfer learning on recurrent architectures}
In this subsection we investigate the effect of the multilingual transfer learning for each model. We first use the rich resource closest language (based on the LID prediction shown in the table) to train  DNN, LSTM and PAC-RNN models, and then adapt them to the target language. The lower part of Table~\ref{tab:results} summarizes the ASR results. As shown, the LSTM models perform significantly better than the baseline SBN system. Using the PAC-RNN model yields a noticeable improvement over the LSTM. Similarly, the PAC-RNN-LSTM can further improve the results. 

\section{Conclusion}
In this paper, we explored a PAC-RNN model for low-resource language speech recognition. The results on multiple languages demonstrated that the PAC-RNN achieves better performance than DNNs and LSTMs. We also showed that by replacing the correction model in the PAC-RNN with an LSTM could further enhance the model. Moreover, the multilingual experiment results show that traditional DNN transfer learning approaches can also be applied to the PAC-RNN architecture. Our future work includes applying the PAC-RNN to tasks on which conventional models do not work well, and extending it by predicting additional information such as the speech signal, speaker, speaking rate, and noise.

\subsubsection*{Acknowledgements}
{\small The authors would like to thank everyone in the Babelon team for feedback and support for various resources. The work uses the following language packs: Cantonese (IARPA-babel101-v0.4c), Turkish (IARPA-babel105b-v0.4), Pashto (IARPA-babel104b-v0.4aY), Tagalog (IARPA-babel106-v0.2g) and Vietnamese (IARPA-babel107b-v0.7),  Assamese (IARPA-babel103b-v0.3), Lao (IARPA-babel203b-v2.1a), Bengali (IARPA-babel102b-v0.4), Zulu  (IARPA-babel206b-v0.1e), Tamil (IARPA-babel204b-v1.1b), Cebuano (IARPA-babel301b-v2.0b), Kurmanji (IARPA-babel205b-v1.0a), and Swahili (IARPA-babel202b-v1.0d-build).}

\bibliographystyle{IEEEbib}
\bibliography{strings,refs}

\end{document}